# HyNeuralMap: Hyperbolic Mapping of Visual Semantics to Neural Hierarchies

Zihan Ma, Tian Xia, Kexin Wang, Xiao Li, Xiaowei He and Yudan Ren*

School of Electronic Information (School of Artificial Intelligence), the Xi'an Key Laboratory of Radiomics and Intelligent Perception, Northwest University, Xi'an 710127, China
{202233494@stumail.nwu.edu.cn, yudan.ren@nwu.edu.cn}

**Abstract.** Understanding the intricate mappings between visual stimuli and neural responses is a fundamental challenge in cognitive neuroscience. While current approaches predominantly align images and functional magnetic resonance imaging (fMRI) responses in Euclidean space, this geometry often struggles to preserve fine-grained semantic relationships and latent hierarchical structures across visual and neural modalities. To overcome this, we propose HyNeuralMap, a framework that employ hyperbolic Lorentz model to map visual semantics into a shared, cross-subject neural hierarchy. By leveraging the negative curvature of hyperbolic space as an inductive bias, the proposed framework better captures hierarchical semantic organization and cross-subject neural similarities. Specifically, visual and neural embeddings are jointly optimized through hyperbolic geometric alignment, where geodesic distances preserve semantic proximity and hierarchical relationships more effectively than Euclidean embeddings. Experiments demonstrate that HyNeuralMap consistently outperforms state-of-the-art Euclidean baselines in both multi-label semantic prediction and cross-modal retrieval tasks. This confirms hyperbolic geometry's superiority for cross-modal semantic alignment and hierarchical modeling, providing a new avenue for vision-neural representation learning.

## 1 Introduction

The brain, as the core of human cognition and perception of the world, constantly encodes the various external stimuli that we encounter daily. Recent advancements have significantly deciphered semantic information from brain responses to visual stimuli [1-7]. These methods utilize functional magnetic resonance imaging (fMRI)-acquired neural patterns to learn meaningful brain features, aligning them with image features from the visual encoders of pretrained vision-language models (VLMs) to unravel how the brain interprets the visual world. However, whether relying on global alignment objectives or modeling hierarchical visual cortical processing, existing methods operate in Euclidean spaces, which struggle to preserve fine-grained semantic relationships across modalities. In particular, semantic embeddings derived from VLM image encoders naturally exhibit structured semantic organization, where coarse concepts (e.g.,

"animal") and fine-grained categories (e.g., "cat") are distributed with different levels of abstraction and semantic proximity rather than uniformly in Euclidean space [8, 9]. Correspondingly, neuroscience evidence suggests neural populations encode semantic information at multiple levels of categorical abstraction, indicating that neural representations may also exhibit structured semantic organization [10, 11]. Preserving such semantic structures is critical for accurate vision-neural alignment, motivating the need for geometry-aware representation learning frameworks.

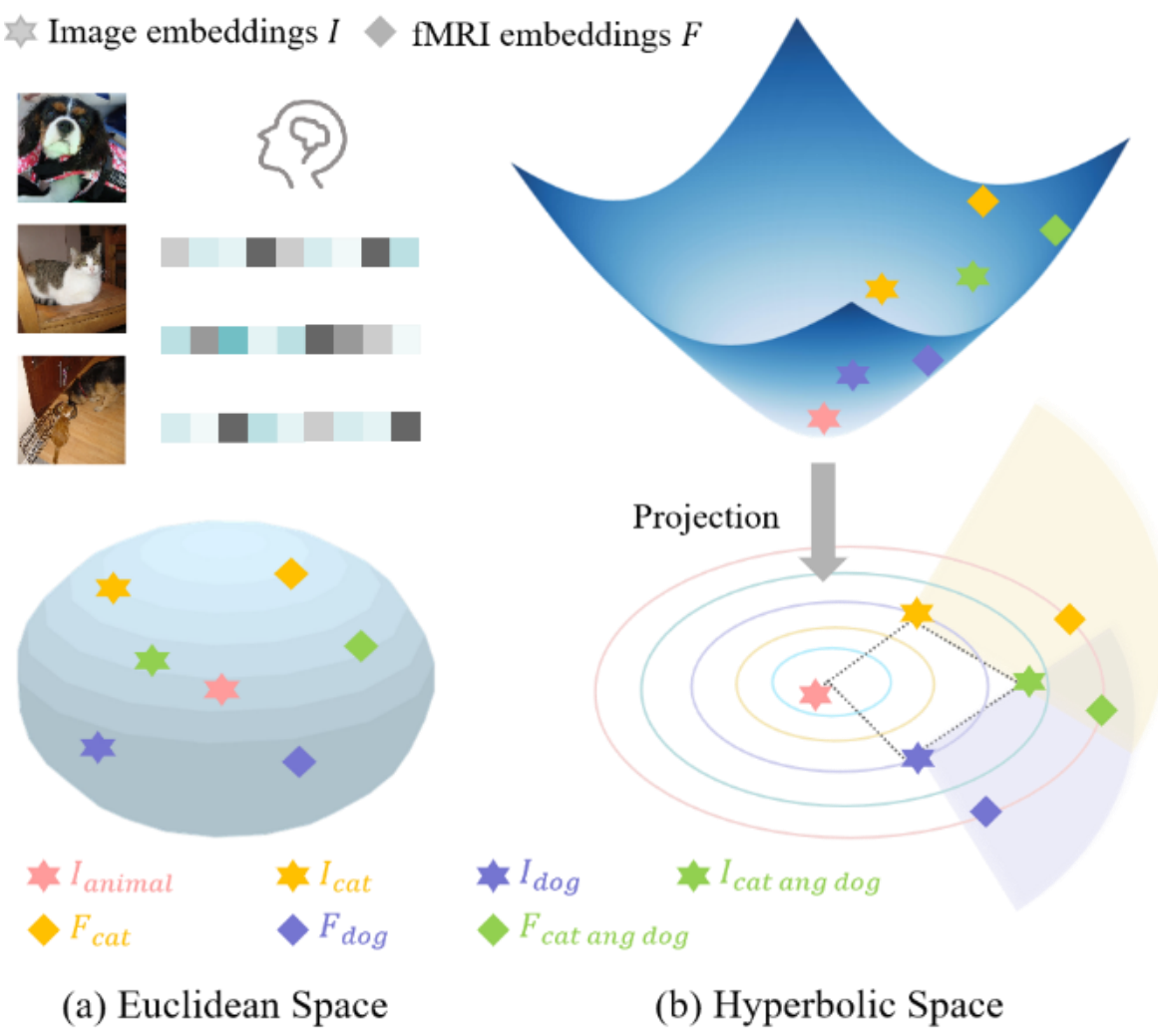

**Fig. 1.** Conceptual comparisons of Euclidean embeddings and hyperbolic embeddings.

However, Euclidean embedding spaces often suffer from semantic crowding when modeling fine-grained semantic relationships, making it difficult to preserve both semantic proximity and abstraction across modalities [12]. As illustrated in Fig. 1(a), Euclidean embeddings primarily reflect semantic similarity but fail to distinguish asymmetric parent-child relationships, such as the ontological link between "animal" and "cat". Conversely, hyperbolic geometry, characterized by its negative curvature and exponential volume growth, provides a rigorous structural inductive bias for modeling the hierarchical nature of vision-neural data [13, 14]. As shown in Fig. 1(b), hyperbolic space inherently organizes information by generality, positioning abstract, broad concepts (e.g., "animal") near the origin, while pushing concrete, fine-grained concepts (e.g., "cat" and "dog") toward the boundary. This geometric arrangement functionally mirrors the "coarse-to-fine" processing strategy of the human visual system, where neural populations initially capture global, coarse semantic categories before iteratively resolving into fine-grained, stimulus-specific details as signal flows through the cortical hierarchy.

Building on this foundation, we establish cross-modal hierarchical constraints for vision-neural alignment based on two key premises. First, we posit that image embeddings from VLMs typically serve as generic semantic prototypes, whereas fMRI responses, as physiological signals triggered by specific visual stimuli, possess significantly higher information specificity. Consequently, for a given concept like "cat", the image embedding $I_{cat}$ is mapped to relatively close to the origin, while the corresponding fMRI embedding $F_{cat}$ resides in a more peripheral region extending radially from $I_{cat}$. Second, this radial arrangement enables the modeling of complex, multi-label stimuli through geometric entailment. For an image containing both "cat" and "dog", the fMRI embedding, $F_{cat\ and\ dog}$ is positioned further from the origin than single-concept embeddings ($F_{cat}$ and $F_{dog}$), specifically residing at the intersection of the "entailment cones" projected by $I_{cat}$ (i.e., the yellow region) and $I_{dog}$ (i.e., the purple region, Fig. 1 (b)). This intersection precisely captures the hierarchical relationship where a composite, specific neural signature is geometrically and semantically entailed by its constituent abstract concepts.

To operationalize this theory, we propose HyNeuralMap, a hyperbolic framework for hierarchical vision-neural alignment. HyNeuralMap leverages hyperbolic geometry to align neural activity with image representations, explicitly modeling their inherent semantic hierarchical structures. It represents concepts from both modalities using the Lorentz model [15]. The learned embeddings are optimized via a novel hyperbolic contrastive loss combined with an entailment loss, capturing both semantic relationships and hierarchical dependencies. Crucially, our approach employs fMRI data from multiple subjects to train a dedicated hyperbolic encoder, enabling the model to learn shared neural response patterns across subjects while retaining subject-specific information, thus addressing a key challenge in cross-subject vision-neural learning.

To validate the effectiveness of the proposed method, we conduct experiments on the Natural Scenes Dataset (NSD) [16]. Results indicate that by preserving the hierarchies of visual representations and neural responses within a shared hyperbolic embedding, our method better captures semantic correlations and hierarchical relationships across modalities, demonstrating significant improvements in semantic decoding tasks.

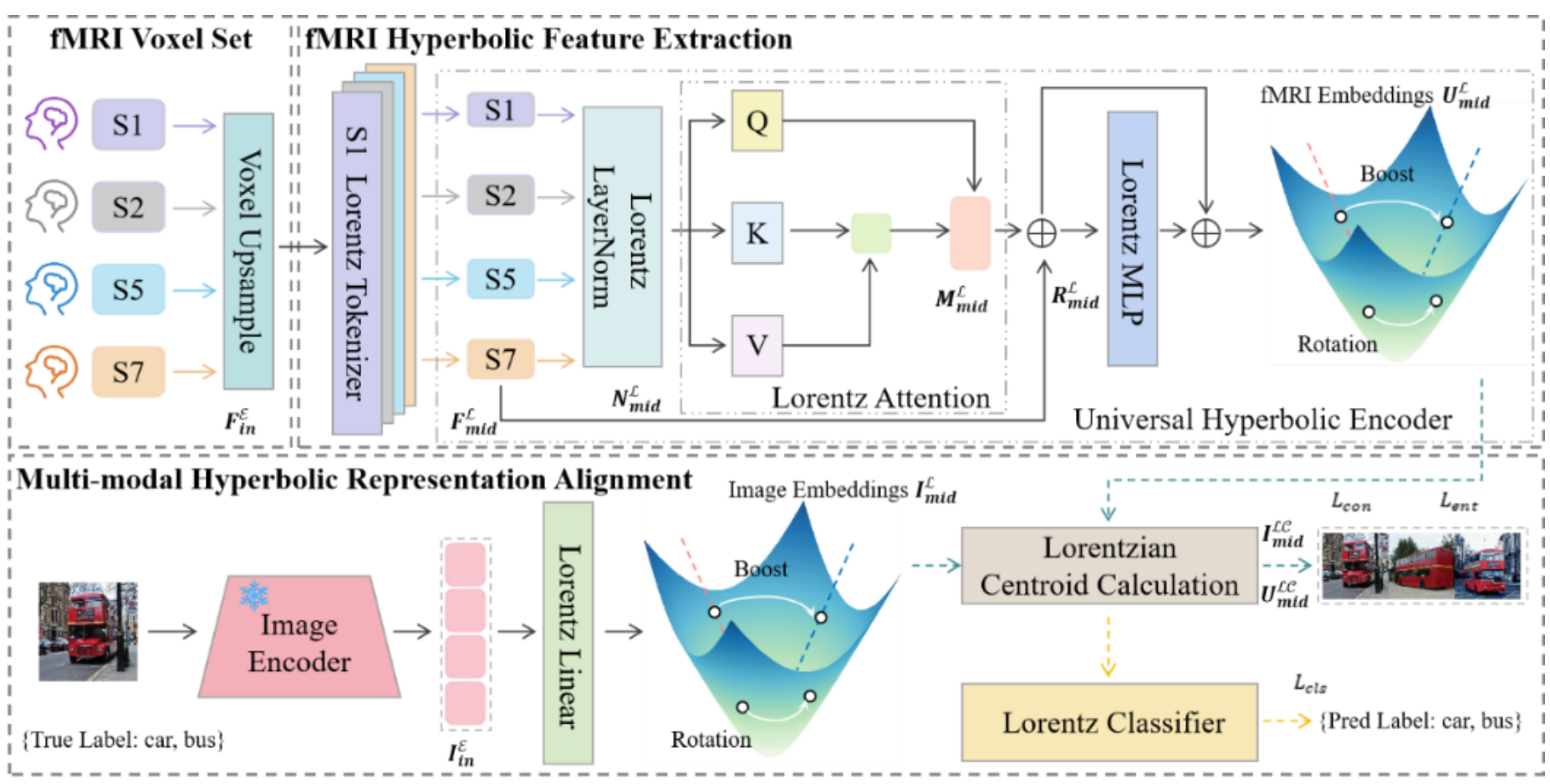

**Fig. 2.** An overview of the proposed HyNeuralMap framework.

# 2 Methodology

We propose a cross-subject vision-neural representation learning scheme, HyNeuralMap, which explicitly captures complex semantic relationships by leveraging the intrinsic hierarchical structure of both fMRI and image data. In Fig. 2, our method aligns cross-modal semantic embeddings in a unified hyperbolic space comprising two key components: fMRI hyperbolic feature extraction and multi-modal hyperbolic representation alignment.

## 2.1 fMRI Hyperbolic Feature Extraction

To address inter-subject brain activity variability [17] and enhance model generalization, we introduce a novel fMRI encoder. This encoder learns hyperbolic embeddings from multi-subject data. The architecture consists of a cross-subject hyperbolic tokenizer, which extracts subject-specific neural representations, and a universal hyperbolic encoder, which captures shared response patterns.

**Cross-subject Hyperbolic Tokenizer.** Initial brain signals $F_{in}^{\mathcal{E}} \in \mathbb{R}^{1\times(v+1)}$, with $v$ as the unified voxel dimension, are projected into a hyperbolic manifold using a Lorentz linear layer. This generates a rich token representation, denoted as $F_{in}^{\mathcal{L}} \in \mathbb{L}^{t}, \mathbb{R}^{(t+1)\times(v+1)}$. To capture subject-specific information, $F_{in}^{\mathcal{L}}$ is then fed into a subject-specific tokenizer, yielding the target embedding $F_{mid}^{\mathcal{L}} \in \mathbb{L}^{d}, \mathbb{R}^{t\times(d+1)}$. Specifically, the spatial component $F_{in-space}^{\mathcal{L}}$ is extracted from $F_{in}^{\mathcal{L}}$, linearly transformed, and concatenated with the recomputed temporal component $F_{in-time}^{\mathcal{L}}$ to form the subject-specific fMRI embedding $F_{mid}^{\mathcal{L}}$. Our operation, diverging from a full Lorentzian transformation [18], is defined as follows:

$$F_{in-time}^{\mathcal{L}} = \sqrt{1/c_{in} + \left\|W^{T}F_{in-space}^{\mathcal{L}} + b\right\|_2^2} \quad (1)$$

$$F_{mid}^{\mathcal{L}} = \left[F_{in-time}^{\mathcal{L}}, W^{T}F_{in-space}^{\mathcal{L}} + b\right] \cdot \sqrt{c_{mid}/c_{in}} \quad (2)$$

where $W$ represents the weight matrix, $b$ is the bias term, and $c_{in}$ and $c_{mid}$ denote the curvature parameters of the input and output manifolds, respectively.

**Universal Hyperbolic Encoder.** The universal hyperbolic encoder maps brain semantic embeddings across subjects into a shared latent space. We employ a Lorentz geometry-based Transformer encoder for fMRI representations.

*Lorentz Layer Normalization.* To maintain the properties of fMRI embedding $F_{mid}^{\mathcal{L}}$ in the hyperbolic manifold, we apply a Lorentz normalization layer [19]. The spatial component $F_{mid-space}^{\mathcal{L}}$ undergoes standard layer normalization and is then concatenated with the temporal component $F_{mid-time}^{\mathcal{L}}$ to obtain the feature, $N_{mid}^{\mathcal{L}}$.

*Lorentz Multi-head Self-attention.* Within the encoder, the Lorentz self-attention mechanism [18] is crucial for capturing intricate fMRI feature dependencies in hyperbolic geometry. Specifically, $N_{mid}^{\mathcal{L}}$ is linearly transformed via learnable weight matrices to generate the query $(Q^{\mathcal{L}})$, key$(K^{\mathcal{L}})$, and value $(V^{\mathcal{L}})$ representations on the Lorentz manifold, a process analogous to the hyperbolic linear transformations described in equations (1) and (2). Attention weights are derived by computing the Lorentzian inner product between $Q^{\mathcal{L}}$ and $K^{\mathcal{L}}$, normalized via the Softmax function, and subsequently used for a weighted summation of $V^{\mathcal{L}}$. This process is delineated as follows:

$$\alpha_{ij} = Softmax(\langle (Q^{\mathcal{L}})^T, (K^{\mathcal{L}})^T \rangle_{\mathbb{L}}) \tag{3}$$

$$Att_i \odot^{c_{mid}} V_j^{\mathcal{L}} := \frac{\sum_{j=1}^{N} \alpha_{ij} V_j^{\mathcal{L}}}{\sqrt{c_{mid} \left| \left\| \sum_{k=1}^{N} \alpha_{ik} V_k^{\mathcal{L}} \right\|_2^2 \right|}} \tag{4}$$

where $\odot^{c_{mid}}$ signifies the weighted sum in the hyperbolic space, and $Att_i$ represents the $i$-th row of the attention matrix within the Lorentz model. After the Lorentz centroid operation, the outputs from all attention heads are concatenated along the feature dimension. A subsequent hyperbolic linear transformation then yields the final feature representation $M_{mid}^{\mathcal{L}} \in \mathbb{L}^d, \mathbb{R}^{t\times(d+1)}$.

*Lorentz Residual Connection.* To facilitate deeper features learning within the model, we incorporate a residual connection [20] between the output of the hyperbolic multi-head self-attention mechanism and the original fMRI embedding $F_{mid}^{\mathcal{L}}$. It is formulated as:

$$R_{mid}^{\mathcal{L}} = \frac{F_{mid}^{\mathcal{L}} + \beta M_{mid}^{\mathcal{L}}}{\sqrt{c_{mid} \left| \langle F_{mid}^{\mathcal{L}} + \beta M_{mid}^{\mathcal{L}}, F_{mid}^{\mathcal{L}} + \beta M_{mid}^{\mathcal{L}} \rangle_{\mathbb{L}} \right|}} \tag{5}$$

Here, $\beta$ balances the weights of the two feature sets.

*Lorentz MLP.* We introduce a Lorentz MLP network to enhance the model's expressive power. This network integrates non-linearity through intrinsic Lorentz linear layers and activation functions [19], which are crucial for extracting richer, higher-level semantic information from the input feature $R_{mid}^{\mathcal{L}}$. Subsequently, the output is residually connected with $R_{mid}^{\mathcal{L}}$, yielding the universal fMRI embedding $U_{mid}^{\mathcal{L}} \in \mathbb{L}^d, \mathbb{R}^{t\times(d+1)}$, which provides a more robust representation for cross-modal learning.

### 2.2 Multi-modal Hyperbolic Representation Alignment

To exploit the VLM-human vision hierarchical alignment [21, 22], we align frozen VLM features with fMRI representations in a hyperbolic space. By employing hyperbolic contrastive and entailment losses, we unify these embeddings to uncover hierarchical vision-neural relationships.

**Lorentzian Centroid Calculation.** Given the extracted image features $I_{mid}^{\mathcal{E}} \in \mathbb{R}^{t\times(p+1)}$, we first transform them from Euclidean geometry to hyperbolic geometry using a Lorentz linear layer, resulting in $I_{mid}^{\mathcal{L}} \in \mathbb{L}^{d}, \mathbb{R}^{t\times(d+1)}$. To extract representative features, we compute the Lorentz centroids $I_{mid}^{\mathcal{LC}}$, $U_{mid}^{\mathcal{LC}} \in \mathbb{R}^{d+1}$ for both $U_{mid}^{\mathcal{L}}$ and $I_{mid}^{\mathcal{L}}$:

$$I_{mid}^{\mathcal{LC}} = \frac{\sum_{j=1}^{N} I_j^{\mathcal{L}}}{\sqrt{c_{mid}\left|\|\sum_{k=1}^{N} I_k^{\mathcal{L}}\|_2^2\right|}} \tag{6}$$

$$U_{mid}^{\mathcal{LC}} = \frac{\sum_{j=1}^{N} U_j^{\mathcal{L}}}{\sqrt{c_{mid}\left|\|\sum_{k=1}^{N} U_k^{\mathcal{L}}\|_2^2\right|}} \tag{7}$$

**Hyperbolic Contrastive Learning.** In cross-modal learning, aligning and comprehending the relationships between diverse modalities frequently employs the contrastive learning paradigm [23]. This study leverages hyperbolic embeddings to align visual data with corresponding brain activity. Given a batch of $N$ samples, we utilize the negative Lorentzian distance as a similarity metric to compute the contrastive loss for image-fMRI data pairs. This objective integrates a learnable temperature parameter $\tau$ and the Softmax function:

$$L_{con}(I,U) = -\sum_{i\in N} log \frac{\exp\left(-d_{\mathbb{L}}\left(I_i^{\mathcal{LC}}, U_i^{\mathcal{LC}}\right)/\tau\right)}{\sum_{k=1.k\neq i}^{N} \exp\left(-d_{\mathbb{L}}\left(I_i^{\mathcal{LC}}, U_k^{\mathcal{LC}}\right)/\tau\right)} \tag{8}$$

Here, for a given image embedding $I_i^{\mathcal{LC}}$, its negative samples are chosen from other fMRI embeddings $U_k^{\mathcal{LC}} (k \neq i)$ within the same batch. Conversely, if an fMRI hyperbolic embedding is the anchor, the loss for negative samples from the batch's image features is defined as $L_{con}(U,I)$. The hyperbolic contrastive loss, which integrates this bidirectional contrastive process, expressed as:

$$L_{con} = \frac{1}{2}\left(L_{con}(I,U) + L_{con}(U,I)\right) \tag{9}$$

This objective promotes the convergence of matched features on the Lorentz manifold, while separating mismatched ones, achieving effective cross-modal alignment.

**Hyperbolic Entailment Learning.** In addition to the contrastive loss, we introduce an entailment loss to reinforce a partial order where the image embedding represents general semantics while the fMRI embedding captures specific neural patterns [24, 25] For notation simplicity, let $x = I_{mid}^{\mathcal{LC}}$ and $y = U_{mid}^{\mathcal{LC}}$. We define an entailment cone [26] for the image embedding $x$ in the hyperbolic space, with its aperture angle given by [27, 28]:

$$aper(x) = \sin^{-1}\left(\frac{2K}{\sqrt{c_{mid}}\|x_{space}\|}\right) \tag{10}$$

where the constant $K = 0.1$ provides stable boundary conditions near the origin. To ensure the concrete concept $y$ resides within the cone of the general concept $x$, we minimize the entailment loss:

$$L_{ent}(x, y) = \max\big(0, ext(x, y) - aper(x)\big) \tag{11}$$

Here, $ext(x, y)$ denotes the external angle of point $y$ with respect to $x$, calculated as:

$$ext(x, y) = \cos^{-1}\left(\frac{y_{time} + x_{time} c_{mid} \langle x, y \rangle_{\mathbb{L}}}{\|x_{space}\| \sqrt{(c_{mid} \langle x, y \rangle_{\mathbb{L}})^2 - 1}}\right) \tag{12}$$

The total loss combines $L_{con}$ and $L_{ent}$:

$$L = L_{con} + \lambda L_{ent} \tag{13}$$

**Lorentz Classifier.** Furthermore, the aggregated neural representation $U_{mid}^{\mathcal{LC}}$ drives multi-label prediction via a Lorentz multinomial logistic regression layer [29], which computes signed hyperbolic distances to learned hyperplanes. Continuing with the notation $y = U_{mid}^{\mathcal{LC}}$, the logits are calculated as:

$$logits(y) = sign(\alpha)\gamma\left(\sqrt{c_{out}}\left|\sinh^{-1}\left(\frac{\alpha}{\gamma\sqrt{c_{out}}}\right)\right|\right) \tag{14}$$

Here, $\alpha$ is the projection of $y$, $\gamma$ is the normal vector's Lorentz norm, and $c_{out}$ corresponds to the output manifold curvature. The classifier is trained via cross-entropy loss.

# 3 Experiments and Results

## 3.1 Dataset and Preprocessing

We conduct experiments on the NSD [16], the most extensive public accessible fMRI dataset, encompassing high-resolution 7-Tesla fMRI scans acquired from eight participants. The visual stimuli presented during fMRI recordings consisted of natural images sourced from the COCO dataset [30], each annotated with multiple labels across 80 categories. For the training set, each participant contributes 8,859 image stimuli and 24,980 fMRI trials. Conversely, the test set for each participant comprises 982 image stimuli and 2,770 fMRI trials. Notably, training images are unique to each participant, whereas test images are consistent across all participants. We primarily utilize Subjects 1, 2, 5, and 7 (S1, S2, S5, S7), who completed the full scanning protocol. Furthermore, owing to inherent variations in brain size and structure [31], the number of voxels varies across participants, typically ranging from approximately 13,000 to 16,000. To facilitate a standardized analysis, these fMRI signals undergo subsequent upsampling, ensuring a uniform voxel count of 18,000 for all participants.

### 3.2 Implementation Details

To evaluate Lorentz representations, we train fMRI encoders in both geometric spaces using CLIP [23], BLIP-2 [32], and DeepSeek-Janus-Pro [33]. The fMRI token count $t$ aligns with the visual dimensions ($257 \times 768$, $257 \times 1408$, and $576 \times 1024$). We project representations onto a 512-d Lorentz manifold. Learnable curvatures $c_{mid}$ and $c_{out}$ are initialized to 1 and 2 (constrained to [0.1,10]), while $c_{in}$ is fixed at 1. Temperature $\tau$ starts at 0.07 (clamped at 0.01), and all scalars are learned in log space.

We employ a hybrid optimization strategy: AdamW for Euclidean parameters and Riemannian Adam for hyperbolic ones, both initialized with a learning rate of 2e-4 and a weight decay of 0.1. Training is performed on an NVIDIA RTX 4090 (batch size 32) for 200 epochs (retrieval) or 100 epochs (classification), with $\lambda$ set to 0.01. For fair comparison, the Euclidean baseline adopts the same Transformer-based architecture and all training configurations listed above, differing only in the underlying geometric space. The additional learnable parameters unique to the Lorentz setting (curvature scalars and temperature scalar) are negligible relative to the shared encoder.

Performance is evaluated using mean Average Precision (mAP) for ranking quality, the area under the receiver operating characteristic curve (AUC) for class discrimination, and Hamming Distance (Ham) to quantify prediction errors.

**Table 1.** Multi-label prediction performance comparison across all subjects and embedding spaces. Bold font signifies the best average performance.

| Manifold | Sub | mAP ↑ | AUC ↑ | Ham ↓ |
|---|---|---|---|---|
| Euclidean | S1 | 0.278 | 0.916 | 0.026 |
| | S2 | 0.268 | 0.916 | 0.027 |
| | S5 | 0.305 | 0.926 | 0.026 |
| | S7 | 0.263 | 0.912 | 0.027 |
| | Avg | 0.279 | 0.917 | 0.027 |
| Hyperbolic | S1 | 0.370 *** | 0.928 * | 0.026 |
| | S2 | 0.345 *** | 0.925 *** | 0.026 |
| | S5 | 0.398 *** | 0.939 ** | 0.026 |
| | S7 | 0.342 * | 0.921 ** | 0.027 |
| | Avg | **0.364** | **0.929** | **0.026** |

Note: * denotes p-value < 0.05; ** denotes p-value < 0.01; *** denotes p-value < 0.001.

### 3.3 Multi-label Prediction

Multi-label prediction aims to decode the brain's semantic representation of specific visual concepts in observed images. Table 1 presents the results of the multi-label prediction task. Demonstrating the effectiveness of the Lorentz manifold, the hyperbolic embeddings learned by HyNeuralMap achieved an 8.5% increase in average mAP compared to Euclidean fMRI embeddings, with the improvement reaching statistical significance for each individual subject. Furthermore, performance across subjects was highly consistent, with a disparity of less than 1.8% in the AUC. This indicates that our model effectively accommodates inter-subject variability, demonstrating strong robustness and generalizability.

### 3.4 Brain-image Retrieval

The retrieval evaluation assesses the amount of image-specific information captured within brain embeddings. We conduct two experiments: image retrieval, which uses a brain embedding to retrieve the most similar image embedding, and brain retrieval, the reverse process. We follow the methodology of Lin et al. [34] to calculate retrieval metrics. Table 2 exports the average Top-1 retrieval accuracy across all subjects for a series of model variants within different embedding spaces. We find that hyperbolic embedding space significantly outperforms Euclidean space with cosine similarity in terms of accuracy. Specifically, Hyperbolic-DeepSeek achieves image retrieval accuracy of 87.8% and brain retrieval accuracy of 87.4%, marking statistically significant improvements of 9.7% and 13.0% over Euclidean-DeepSeek. This suggests that hyperbolic space is more effective at capturing the inherent hierarchical structures and complex relationships within fMRI and image data, thereby reducing the semantic gap between different modalities. Notably, the advantages of hyperbolic space extend to all three variants, underscoring the broad applicability of our framework.

**Table 2.** Retrieval performance comparison across Euclidean and Hyperbolic spaces and VLMs. All results are averaged across 4 subjects. Bold font signifies the best avg. performance.

| Manifold | VLMs | Image ↑ | Brain ↑ |
|---|---|---|---|
| Euclidean | CLIP | 62.3% | 70.5% |
| | BLIP-2 | 82.9% | 71.0% |
| | DeepSeek | 78.1% | 74.4% |
| Hyperbolic | CLIP | 76.2% *** | 76.3% *** |
| | BLIP-2 | 85.3% | 84.9% *** |
| | DeepSeek | **87.8%** *** | **87.4%** *** |

Note: * denotes p-value < 0.05; ** denotes p-value < 0.01; *** denotes p-value < 0.001.

### 3.5 Ablation Studies

**Efficacy of Architectural Design.** We first investigate the impact of fMRI encoder architectures on model performance. As shown in Table 3, our method consistently outperforms the MLP-based architecture in both multi-label prediction and retrieval tasks. This demonstrates that the Lorentz Transformer more adeptly captures complex nonlinear relationships within hyperbolic fMRI data, enhancing the model's expressive power.

**Efficacy of the Entailment Loss.** We analyze the effectiveness of the entailment loss in guiding model learning, as detailed in Table 3. Removing the entailment loss (denoted as w/o $L_{ent}$) leads to a drop in image retrieval accuracy from 87.8% (for our HyNeuralMap model) to 87.1%, indicating its beneficial role in vision-neural mapping. The complete model consistently achieves superior performance across both metrics, emphasizing the necessity of synergistic contributions from multiple loss functions to enhance model performance.

**Impact of Lorentz Curvature**. In the Lorentz model, the setting of curvature is crucial for effective embedding representation learning. To identify the optimal configuration, we compare the performance impact of a fixed versus a learnable $c_{mid}$ parameter. Table 3 demonstrates that treating curvature as a learnable parameter consistently achieves superior results in both tasks.

**Table 3.** Performance comparison of ablation study with DeepSeek backbone. All results are averaged across 4 subjects. Bold font signifies the best performance.

| Method | Multi-label prediction | | | Retrieval | |
|---|---|---|---|---|---|
| | mAP ↑ | AUC ↑ | Ham ↓ | Image ↑ | Brain ↑ |
| Efficacy of Architectural Design | | | | | |
| MLP | 0.322 | 0.911 | 0.026 | 85.7% | 84.5% |
| Effectiveness of the Entailment Loss | | | | | |
| w/o $L_{ent}$ | - | - | - | 87.1% | 87.3% |
| Impact of Lorentz Curvature | | | | | |
| 1.0 | 0.332 | 0.915 | 0.027 | 87.7% | 87.3% |
| 2.0 | 0.310 | 0.913 | 0.029 | 87.7% | 83.4% |
| 3.0 | 0.248 | 0.893 | 0.029 | 86.7% | 83.2% |
| Our | **0.364** | **0.929** | **0.026** | **87.8%** | **87.4%** |

### 3.6 Visualization of Hyperbolic Space

To explore the semantic distribution of fMRI and image embeddings, we visually analyze the learned hyperbolic space. First, we examine the global distribution by randomly sampling 1K training embeddings for norm histogram analysis and employ CO-SNE [35] for low-dimensional projection. Fig. 3 (a) clearly shows that image embeddings are positioned closer to the center of the hyperbolic space compared to fMRI embeddings. This radial shift empirically validates our strategy of enforcing a partial order relationship, confirming that the model treats neural responses as high-specificity descendants of abstract visual prototypes. Furthermore, Fig. 3 (b) reveals distinct semantic separation and hierarchical structures within the embedding distribution, indicating that HyNeuralMap successfully captures latent semantic associations and macroscopic hierarchical relationships across modalities.

Further, to verify the fine-grained semantic structure at the instance level, we visualize representative hierarchical chains from the test set in Fig. 4. The single-chain structure (left) confirm that the model encodes strict "parent-child" hierarchical relationships via hyperbolic radii. Taking the "electronics" category as an example, the parent visual prototype resides near the origin ($r = 0.14$), acting as a semantic root node, while the subclass "cell phone" is distributed in the outer layer ($r = 0.291$). This significant radial increase indicates that our method constructs an effective entailment cone, correctly reflecting the hierarchy from abstract to specific. Moreover, this hierarchy also extends to the neural domain: the fMRI embedding for "cell phone" is located at a deeper hyperbolic radius ($r = 0.468$), adhering to the learned cross-modal partial order.

The multi-branch structure (Fig. 4, right) further highlights the model's capacity to handle complex, composite concepts. Within the "food" cluster, although "apple" ($r =$ 0.309) and "carrot" ($r = 0.303$) belong to different subclasses, they maintain similar radial depths, reflecting their status as sibling nodes, while achieving clear separation in the angular space. Notably, for the composite image "apple and carrot", the image embedding naturally resides in the region between the two subclasses, whereas its corresponding fMRI signal extends further toward the boundary, accurately falling into the intersection of the entailment cones formed by the "apple" and "carrot" visual embeddings. This provides strong evidence that the model precisely captures the hierarchical entailment structure of cross-modal semantic concepts.

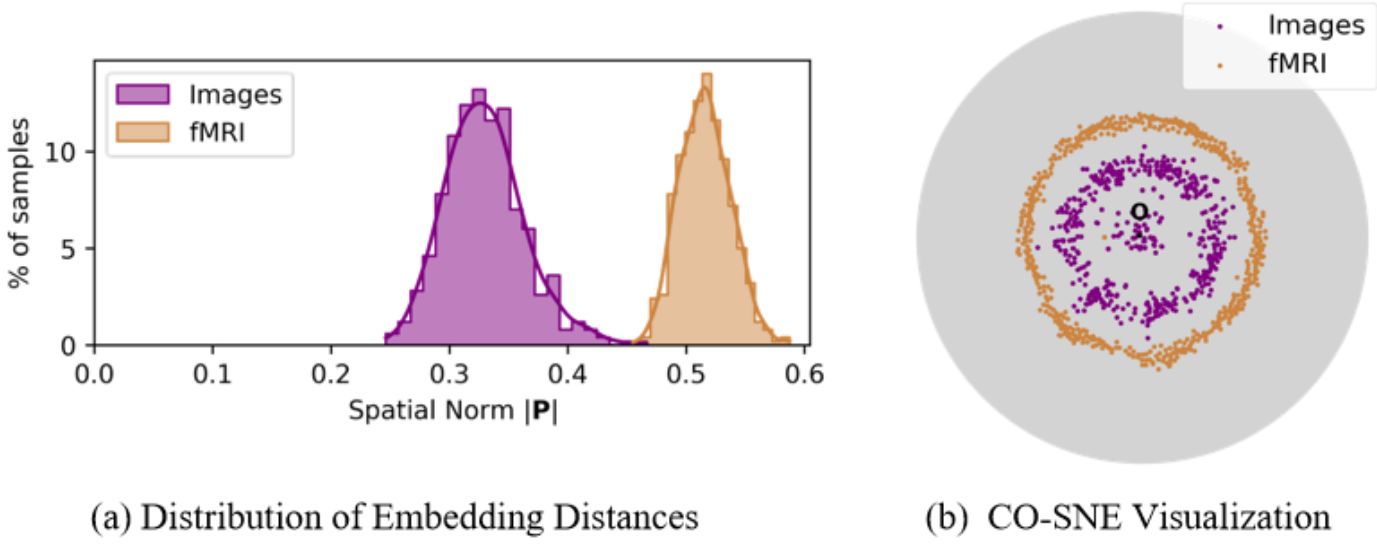

(a) Distribution of Embedding Distances (b) CO-SNE Visualization

**Fig. 3.** Visualization of the learned hyperbolic space by HyNeuralMap-DeepSeek. Elements closer to the origin exhibit higher semantic hierarchy and coarser granularity.

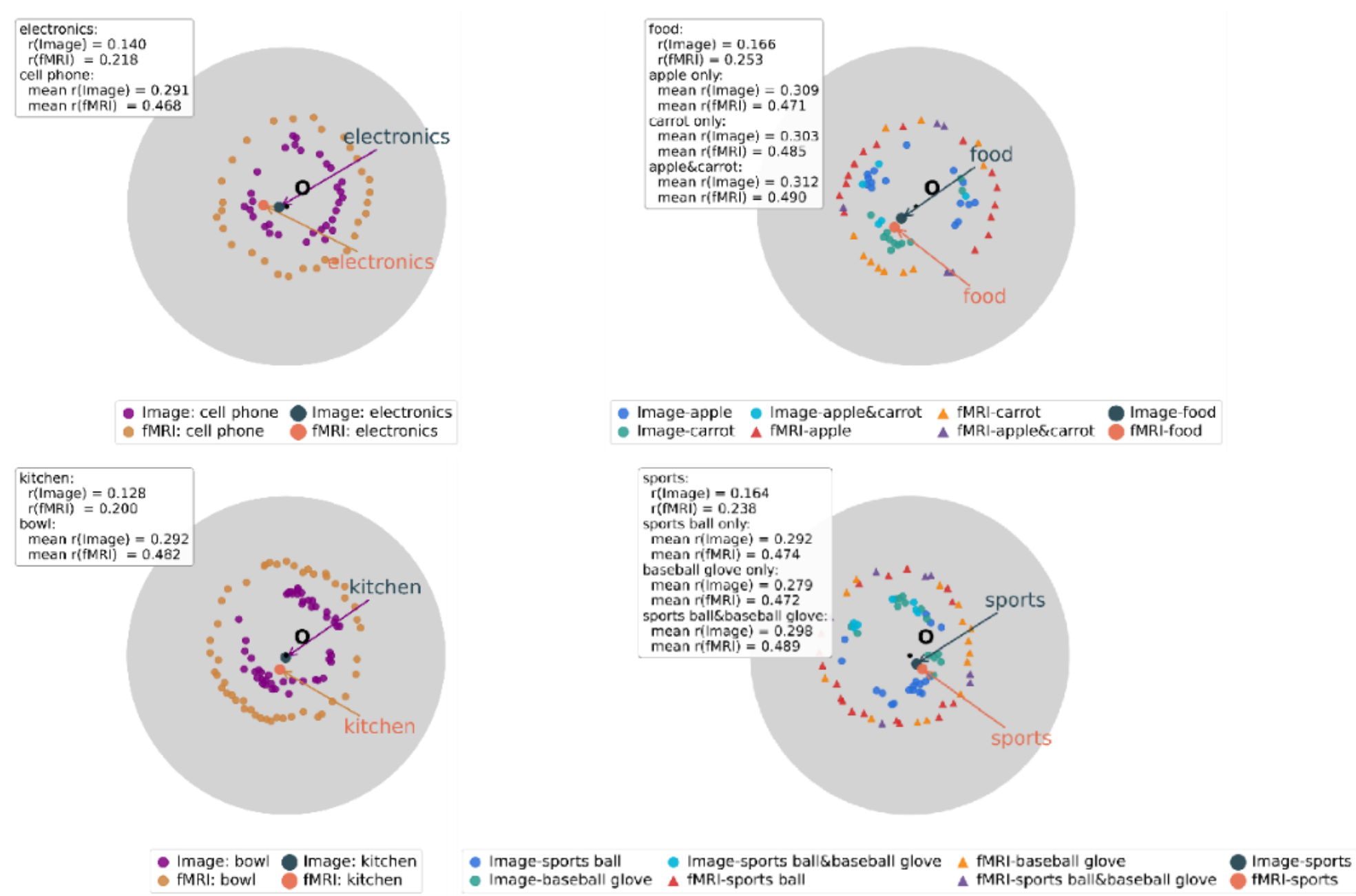

**Fig. 4.** Fine-grained visualization of the hierarchy and semantic compositionality in the hypebolic space.

# 4 Conclusion

In this paper, we introduce HyNeuralMap, a novel hyperbolic framework designed to map visual semantics to neural hierarchies. By aligning neural responses and image representations in a shared Lorentz manifold, our method effectively captures cross-modal semantic hierarchical relationships. We design a hyperbolic fMRI encoder that extracts both shared and individual-specific brain patterns. To achieve robust semantic alignment, we incorporate a novel optimization strategy combining hyperbolic contrastive loss and partial order entailment constraints, yielding more discriminative hyperbolic embeddings. Experiments demonstrate that HyNeuralMap surpasses Euclidean-based approaches. Our visualization analyses further confirm HyNeuralMap's capability to geometrically embed multi-modal hierarchical structures in hyperbolic space. Our work not only opens a new perspective for vision-neural alignment but also underscores the immense potential of hyperbolic geometry in modeling complex cross-modal semantic relationships.